\documentclass[runningheads]{llncs}
\usepackage[T1]{fontenc}
\usepackage{hyperref}
\usepackage{color}

\usepackage{graphicx}
\usepackage{tabularx}
\usepackage{booktabs}
\usepackage{multirow}
\usepackage{algorithmic}
\usepackage{algorithm}
\usepackage{amsmath, amssymb}
\usepackage{bbm}
\usepackage{multicol}
\hypersetup{colorlinks=true,urlcolor=blue, citecolor = blue}
\usepackage{orcidlink}
\usepackage{fontawesome}
\usepackage[symbol]{footmisc}

\begin{document}
\title{FeSViBS: Federated Split Learning of Vision Transformer with Block Sampling}
\titlerunning{FeSViBS}
%
\author{Faris Almalik\textsuperscript{\dag}\orcidlink{0000-0002-7885-6285} \and Naif Alkhunaizi\textsuperscript{\dag}\orcidlink{0000-0002-7093-5034} \and Ibrahim Almakky\orcidlink{0009-0008-8802-7107} \and Karthik Nandakumar\orcidlink{0000-0002-6274-9725}\textsuperscript{(\faEnvelopeO)}}
\authorrunning{F. Almalik et al.}

\institute{Mohamed Bin Zayed University of Artificial Intelligence, Abu Dhabi, UAE
\email{\{faris.almalik, naif.alkhunaizi, ibrahim.almakky, karthik.nandakumar\}@mbzuai.ac.ae}}

\maketitle              
\begin{abstract}
Data scarcity is a significant obstacle hindering the learning of powerful machine learning models in critical healthcare applications. Data-sharing mechanisms among multiple entities (e.g., hospitals) can accelerate model training and yield more accurate predictions. Recently, approaches such as Federated Learning (FL) and Split Learning (SL) have facilitated collaboration without the need to exchange private data. In this work, we propose a framework for medical imaging classification tasks called \textbf{Fe}derated \textbf{S}plit learning of \textbf{Vi}sion transformer with \textbf{B}lock \textbf{S}ampling \textbf{(FeSViBS)}. The FeSViBS framework builds upon the existing federated split vision transformer and introduces a \emph{block sampling} module, which leverages intermediate features extracted by the Vision Transformer (ViT) at the server. This is achieved by sampling features (patch tokens) from an intermediate transformer block and distilling their information content into a pseudo class token before passing them back to the client. These pseudo class tokens serve as an effective feature augmentation strategy and enhances the generalizability of the learned model. We demonstrate the utility of our proposed method compared to other SL and FL approaches on three publicly available medical imaging datasets: HAM1000, BloodMNIST, and Fed-ISIC2019, under both IID and non-IID settings. Code: \url{https://github.com/faresmalik/FeSViBS} 

\end{abstract}

\footnotetext[2]{Equal contribution}

\begin{figure}[t!]
    \centering{    \includegraphics[width=0.9\textwidth]{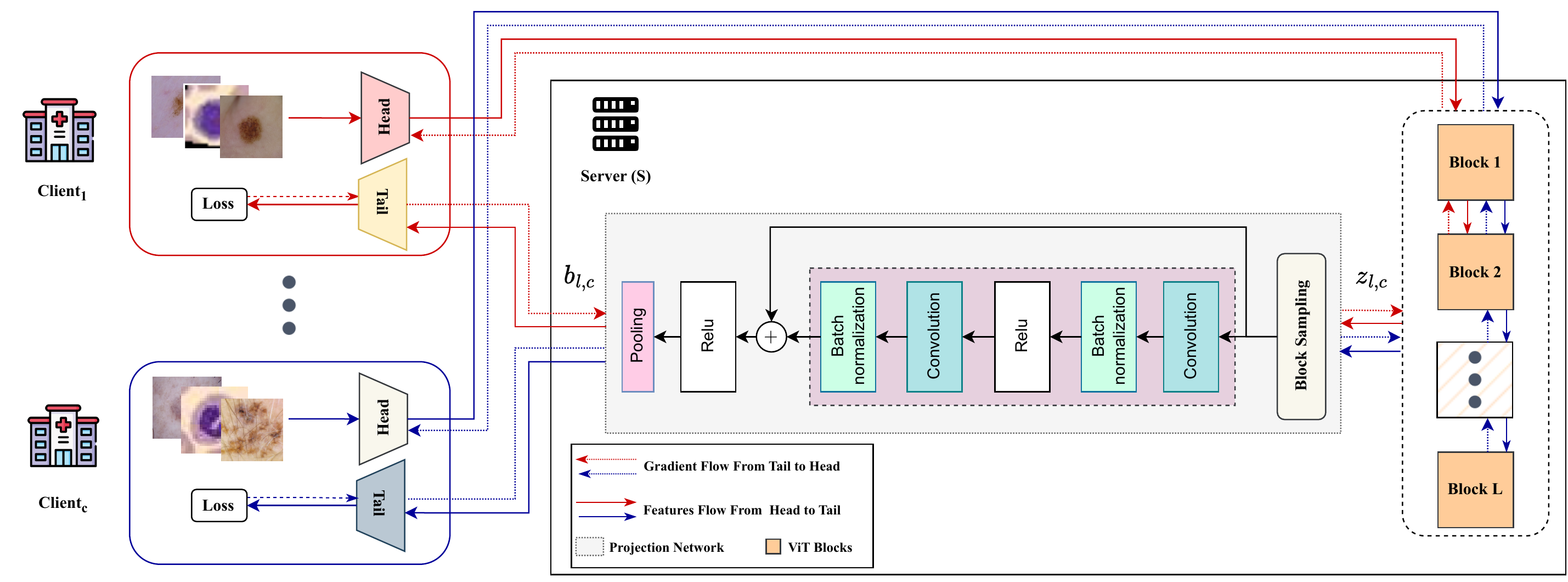}
    }
    \caption{FeSViBS framework. The server receives smashed representations from the clients, samples a ViT block for each client, uses a projection network to distill patch tokens into pseudo class tokens, which are sent back to the client for final prediction.}
    \label{Figure: Method}
\end{figure}

\section{Introduction}
Vision Transformers (ViTs) are self-attention based neural networks that have achieved state-of-the-art performance on various medical imaging tasks \cite{shamshad2022transformers,zheng2021rethinking,ClassificationTransf}. Since ViTs are capable of encoding long range dependencies between input sequences \cite{khan2021transformers}, they are more robust against distribution shifts and are well-suited for handling heterogeneous distributions \cite{bhojanapalli2021understanding}. However, training ViT models typically requires significantly more data than traditional Convolutional Neural Network (CNN) models \cite{khan2021transformers}, which limits their application in domains such as healthcare, where data scarcity is a challenge. One way to overcome this challenge is to train such models in a collaborative and distributed manner, where large amounts of data can be leveraged from different sites without the need for sharing private data \cite{intronew,ha2022feasibility}. Federated learning and split learning are two well-known approaches for collaborative model training.

Federated Learning (FL) enables clients to collaboratively learn a global model by aggregating locally trained models \cite{federatedbook}. Since this can be accomplished without sharing raw data, FL mitigates risks related to private data leakage. Several aggregation rules such as FedAvg \cite{mcmahan2017communication} and FedProx \cite{li2020federated} have been proposed for FL. However, it has been demonstrated that most FL algorithms are vulnerable to gradient inversion attacks \cite{gradientinversion}, which dilute their privacy guarantees. In contrast, Split Learning (SL) divides a deep neural network into components with independently accessible parameters \cite{SplitFederatedGupta}. Since  no participant in SL can access the complete model parameters, it has been claimed that SL offers better data confidentiality compared to FL. In particular, the U-shaped SL configuration, where each client has its own feature extraction head and a task-specific tail \cite{vepakomma2018split} can further improve client privacy, as it circumvents the need to share the data or labels. Recently, SL frameworks have been proposed for various medical applications such as tumor classification \cite{ads2021multi} and chext x-ray classification \cite{splitLearningHealthcare}.

Recent studies \cite{oh2022differentially,visonVits} have demonstrated that both FL and SL can be combined to effectively train ViTs. In \cite{visonVits}, a framework called FeSTA was proposed for medical image classification. The FeSTA framework involves a hybrid ViT architecture with U-shaped SL configuration - each client has its own CNN head and a multilayer perceptron (MLP) tail, while the shared ViT body resides on a central server. This architecture can be trained using both SL and FL in a potentially task-agnostic fashion, leading to better performance compared to other distributed learning methods. The work in \cite{oh2022differentially} focuses on privacy and incorporates differential privacy with mixed masked patches sent from the ViT on the server to the clients to prevent any potential data leakage.

In this work, we build upon on the FeSTA framework \cite{visonVits} for collaborative learning of ViT. Despite its success, FeSTA requires pretraining the ViT body on a large dataset prior to its utilization in the SL and FL training process. In the absence of pretraining, limited training data availability (a common problem in medical imaging) leads to severe overfitting and poor generalization. Furthermore, the FeSTA framework exploits only the final \emph{cls} token produced by the ViT body and ignores all the other intermediate features of the ViT. It is well-known that intermediate features (referred to as patch tokens) also contain discriminative information that could be useful for the classification task \cite{farispaper}. 

To overcome the above limitations, we propose a framework called \textbf{Fe}derated \textbf{S}plit learning of \textbf{Vi}sion transformer with \textbf{B}lock \textbf{S}ampling \textbf{(FeSViBS)}. Our primary novelty is the introduction of a \emph{block sampling} module, which randomly selects an intermediate transformer block for each client in each training round, extracts intermediate features, and distills these features into a pseudo cls token using a shared projection network. The proposed approach has two key benefits: (i) it effectively leverages intermediate ViT features, which are completely ignored in FeSTA, and (ii) sampling these intermediate features from different blocks, rather than relying solely on an individual block's features or the final \emph{cls} token, serves as a feature augmentation strategy for the network, enhancing its generalization. The contributions of this work can be summarized as follows:

\begin{itemize}
    \item [i.] We propose the FeSViBS framework, a novel federated and split learning framework that leverages the features learned by intermediate ViT blocks to enhance the performance of the collaborative system. 
    \item [ii.] We introduce block sampling at the server level, which acts as a feature augmentation strategy for better generalization. 
\end{itemize}

\section{Methodology}
We first describe the working of a typical split vision transformer before proceeding to describe FeSViBS. Each client $c \in [1,n]$ has access to local private data $(x_c, y_c) \in \{x_c^{(i)},y_c^{(i)}\}_{i=1}^{N_c}$, where $N_c$ is the number of training samples available at client $c$, $x$ represents the input data, and $y$ is the class label. Following \cite{visonVits}, we assume U-shaped split learning setting, with each client having two local networks called \emph{head} ($\mathbf{\mathcal{H}}_{\theta_c}$) and \emph{tail} ($\mathbf{\mathcal{T}}_{\psi_c}$), where 
$\theta_c$ and $\psi_c$ are client-specific \emph{head} and \emph{tail} parameters, respectively. The server consists of a ViT \emph{body} ($\mathbf{\mathcal{B}}_\Phi$), which includes a stack of $L$ transformer blocks denoted as $\mathbf{\mathcal{B}}_{\Phi_1}, \mathbf{\mathcal{B}}_{\Phi_2}, \cdots, \mathbf{\mathcal{B}}_{\Phi_L}$ and $\mathbf{\mathcal{B}}_\Phi(\cdot) = \mathbf{\mathcal{B}}_{\Phi_L}(\cdots(\mathbf{\mathcal{B}}_{\Phi_2}(\mathbf{\mathcal{B}}_{\Phi_1}(\cdot))))$. Here, $\Phi_l$ represents the parameters of the $l^{th}$ transformer block and $\Phi = \left[\Phi_1,\Phi_2,\cdots,\Phi_L\right]$ denotes the complete set of parameters of the transformer body.

During training, the client performs a forward pass of the input data through the head to produce an embedding $h_c = \mathbf{\mathcal{H}}_{\theta_c}(x_c) \in \mathbb{R}^{768\times M}$ of its local data, which is typically organized as $M$ \emph{patch tokens} representing different patches of the input image. These embeddings (\emph{smashed} representations) are then sent to the server. The ViT appends an additional token called the class token (\emph{cls} $\in \mathbb{R}^{768\times 1}$) and utilizes the self-attention mechanism to obtain a representation $b_c = \mathbf{\mathcal{B}}_\Phi(h_c) \in \mathbb{R}^{768\times 1}$, which is typically the \emph{cls} token resulting from the last transformer block. This \emph{cls} token is returned to the client for further processing. The \emph{tail} at each client projects the received class token representation $b_c$ into a class probability distribution to get the final prediction $\hat{y}_c = \mathbf{\mathcal{T}}_{\psi_c}(b_c)$. This marks the end of the forward pass. Subsequently, the backpropagation starts with computing loss $\ell_c(y_c,\hat{y}_c)$, where $\ell_c(.)$ represents the client's loss function between the true labels $y_c$ and predicted labels $\hat{y}_c$. The gradient of this loss is propagated back in the reverse order from the client's \emph{tail}, server's \emph{body}, to the client's \emph{head}. We refer to this setting as Split Learning of Vision Transformer (\textbf{SLViT}), where each client optimizes the following objective in each round: 

\begin{equation}
    \min_{\theta_c, \Phi, \psi_{c}} \frac{1}{N_c} \sum_{i=1}^{N_c} \ell_c \big(y_c^{(i)}, \mathbf{\mathcal{T}}_{\psi_c}(\mathbf{\mathcal{B}}_\Phi(\mathbf{\mathcal{H}}_{\theta_c}(x_c^{(i)}))\big)
    \label{eqn:vslobj}
\end{equation}

In FeSTA \cite{visonVits}, an additional federation step was introduced. After every few SL rounds, the local (client-specific) \emph{heads} and \emph{tails} are aggregated in a \emph{unifying round} using FedAvg \cite{mcmahan2017communication} to produce global parameters $\bar{\theta}$ and $\bar{\psi}$. Note that the above framework completely ignores all the intermediate features (\emph{patch} tokens) extracted from various ViT blocks. In \cite{farispaper}, it was demonstrated that these patch tokens are also discriminative and valuable for classification tasks. Hence, we aim to exploit these intermediate features to further enhance the performance.

\subsection{FeSViBS Framework}
The proposed FeSViBS method is illustrated in Fig. \ref{Figure: Method} and detailed in Alg. \ref{alg:cap}. The working of the FeSViBS framework is very similar to FeSTA, except for one key difference. During the forward pass of SLViT and FeSTA, the server always returns the \emph{cls} token from the last ViT block. In contrast, a FeSViBS server samples an intermediate block $l \in \{1,2.\dots,L\}$ for each client $c$ in each round and extracts the intermediate features $z_{c,l}$ from the chosen $l^{th}$ block as follows: 

\begin{equation}
    z_{c,l} = \mathbf{\mathcal{B}}_{\Phi_l} \Big( \mathbf{\mathcal{B}}_{\Phi_{l-1}} \dots \mathbf{\mathcal{B}}_{\Phi_{1}}\big(\mathbf{\mathcal{H}}_{\theta_c}(x_c)\big)\Big)
\end{equation}

\noindent where  $z_{c,l} \in \mathbb{R}^{768 \times M}$. The server then projects the extracted intermediate features into a lower dimension using a \emph{projection network} $\mathcal{R}$ (shared across all blocks) to obtain the final representation $b_{c,l} = \mathcal{R}_{\pi}(z_{c,l})$, where $b_{c,l} \in \mathbb{R}^{768 \times 1}$. This final representation $b_{c,l}$ can be considered as a \emph{pseudo class token} and the role of the projection network is to distill the discriminative information contained in the intermediate features into this pseudo class token. The architecture of the projection network is shown in Figure \ref{Figure: Method} and it resembles a simple ResNet \cite{ResNet} block with skip connection. The pseudo class token is then sent to the client's \emph{tail} to obtain the final prediction $\hat{y}_c = \mathcal{T}_{\psi_c} (b_{c,l})$ and complete the forward pass.

Each client uses $\hat{y}_c$ along with the true labels $y_c$ to compute the loss $\ell_c(y_c, \hat{y}_c)$. The gradients of the client's \emph{tail} are then calculated and sent back to the server, which then carries out the back-propagation through the \emph{projection network} and relevant blocks of the ViT \emph{body} (only those blocks involved in the corresponding forward pass). Next, the server sends the gradients back to the client to propagate them through the \emph{head} and end the back-propagation step. Hence, the client's optimization problem can be formulated as follows: 
\begin{equation}
\label{our_equation}
    \min_{\theta_c, \Phi_{1:l, c}, \pi, \psi_{c}} \frac{1}{N_c}\sum_{i=1}^{N_c} \ell_c \big( y_c^{(i)}, \mathbf{\mathcal{T}}_{\psi_c}(b_{c,l}^{(i)})\big).
\end{equation}

In the FeSViBS framework, the \emph{heads} and \emph{tails} of all the clients are assumed to have the same network architecture. Within each collaboration round,  all the clients perform the forward and backward passes. While the parameters of the relevant head and tail as well as the shared projection network are updated after every backward pass, the parameters of the ViT body are updated only at the end of a collaboration round after aggregating updates from all the clients. The above protocol until this step is referred as \textbf{SViBS} because there is still no federation of the \emph{heads} and \emph{tails}. Similar to FeSTA, we also perform aggregation of the local \emph{heads} and \emph{tails} periodically in unifying rounds, resulting in the final FeSViBS framework. While in SViBS, the clients can initialize their heads and tails independently, FeSviBS requires a common initialization by the server and sharing of aggregated head and tail parameters after a unifying round.

\begin{algorithm}[t!]
    \caption{FeSViBS}
    \label{alg:cap}
    \begin{algorithmic}[1]
    \REQUIRE Local data at client c ($x_c, y_c$). Server initializes the body parameters ($\Phi$), Projection Network parameters ($\pi$), client head and tail parameters ($\bar{\theta}, \bar{\psi}$)
    \FOR {rounds $r =1,2,\dots, R$}
        \FOR {\textbf{client} {$c \in [1,n]$} }
        \IF {$r = 1$ \OR $(r-1) \in \text{Unifying Rounds}$}
        \STATE ($\theta_c, \psi_c$) $\leftarrow$ ($\bar{\theta}, \bar{\psi}$) 
        \ENDIF
        \STATE  \underline{Client c}: \: $h_c\leftarrow \mathbf{\mathcal{H}}_{\theta_c} (x_c)$ 
        \STATE \underline{Server}:
        \STATE $\qquad$ Sample a ViT block ($l$) for client c
        \STATE $\qquad$ $b_{c,l} \leftarrow  \mathcal{R}_{\pi} \Big(\mathbf{\mathcal{B}}_{\Phi_l} \big( \mathbf{\mathcal{B}}_{\Phi_{l-1}} \dots \mathbf{\mathcal{B}}_{\Phi_{1}}(h_c)\big)\Big)$
        \STATE \underline{Client c}:
        \STATE $\qquad$ Compute $\ell_c \big(y_c, \mathbf{\mathcal{T}}_{\psi_c}(b_{c,l})\big)$ and Backprop.
        \STATE $\qquad$ Update ($\theta_c, \psi_c$) with suitable optimizer
        \STATE \underline{Server}:
        \STATE $\qquad$ Update $(\pi)$ with suitable optimizer, Compute and store $\Phi_{1:l, c}$
    \ENDFOR
    \STATE \underline{Server}:
    \STATE $\qquad$ Update body: $\Phi \leftarrow \frac{1}{n} \sum_{c} \Phi_{1:l, c}$ \\
    \IF{$ r \in \text{Unifying Rounds}$}
        \STATE $\qquad$ ($\bar{\theta}, \bar{\psi}$) $\leftarrow$ ($\frac{1}{n}\sum_{c}\theta_c, \frac{1}{n}\sum_{c}\psi_c$)
    \ENDIF
  \ENDFOR
\end{algorithmic}
\end{algorithm}

\section{Experimental Setup}

\begin{figure}[t!]
\centering{
\includegraphics[width=0.95\textwidth]{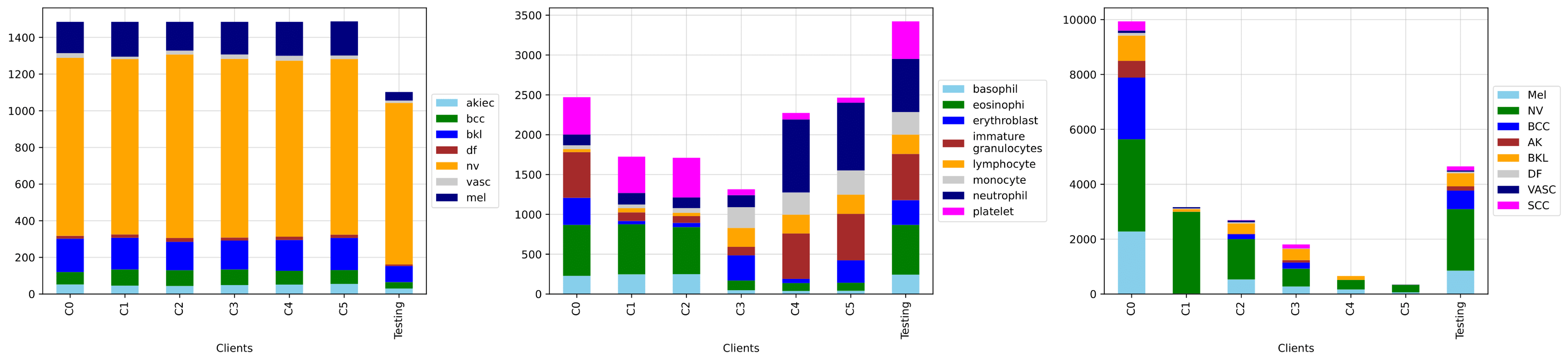}
}
\caption{Distribution of: (\textbf{left}) HAM10000, (\textbf{middle}) BloodMNIST, and (\textbf{right}) Fed-ISIC2019. Each stacked bar represents the number of samples, and each color represents each class. The last bar in each figure represents the testing set.} 
\label{Figure: distributions}
\end{figure}

\noindent \textbf{Datasets.} We conduct our experiments on three medical imaging datasets. The first dataset is HAM10000 \cite{HAM1000}, a multi-class dataset comprising of $10,015$ dermoscopic images from diverse populations. HAM10000 includes $7$ imbalanced categories of pigmented lesions; we randomly perform $80\%/20\%$ split for training and testing, respectively. The second dataset \cite{bloodmnist} termed "BloodMNIST" is a multi-class dataset consisting of $17,092$ blood cell images for $8$ different imbalanced cell types. We followed \cite{medmnist} and split the dataset into $70\%$ training, $10\%$ validation, and $20\%$ testing. Finally, the Fed-ISIC2019 dataset consists of $23,247$ dermoscopy images collected from $6$ centers, with significant differences in population characteristics and acquisition systems, for $8$ different melanoma classes. This dataset was prepared by FLamby \cite{flamby} from the original ISIC2019 \cite{HAM1000,HAM2,HAM3}, where they split the data according to the image acquisition used, generating real federated settings with a natural split across $6$ clients with $80\%/20\%$ split for training and testing, respectively. The training samples in all datasets are divided among $6$ clients, whereas the testing set is shared among them all. The distribution of each dataset is depicted in Fig. \ref{Figure: distributions}. Note that Fed-ISIC2019 and BloodMNIST are non-IID, whereas HAM10000 is IID. 

\noindent \textbf{Server's Network.} For the server's body, we chose the ViT-B/16 model from timm library \cite{timm} which includes $L = 12$ transformer blocks, embedding dimension $D = 768$, $12$ attention heads, and divides the input image into patches each of size $16 \times 16$ with $M = 196$ patches. We limit the block sampling to the first $6$ ViT blocks. Additionally, the projection network has two convolution layers with a skip connection, which takes an input of dimension $768 \times 196$ and projects it into a lower dimension of $768$. 

\noindent \textbf{Clients' Networks.} Each client has two main networks: head and tail. We followed timm's library implementation of Hybrid ViTs to design each client's head and tail. The head network is a ResNet-50 \cite{ResNet} with a convolution layer added to project the features extracted by ResNet-50 to a dimension of $768 \times 196$, while the tail is a classification head. Also, we unify the clients' networks (head and tail) every $2$ rounds using FedAvg. We conduct our experiments for $200$ rounds with Adam optimizer \cite{Adam}, a learning rate of $1\times10^{-4}$, and $32$ batch size with a cross-entropy loss calculated at the tail. The code was implemented using PyTorch $1.10$ and the models were trained using Nvidia A100 GPU with 40 GB memory.

\begin{table}[t!]
\centering
\caption{Performance of different methods. We report the average balanced accuracy across clients with standard deviation for local, SLViT, FeSTA, SViBS, and FeSViBS. Centralized, FedAvg, FedProx, SCAFFOLD and MOON has one global unified model for all clients.}
\label{tab:my-table}
\begin{tabular}{@{}lccccc@{}}
\toprule
\multirow{2}{*}{} & \multicolumn{5}{c}{\textbf{Dataset}}                                       \\ \cmidrule(l){2-6} 
                  & HAM10000 &  & BloodMNIST &  & Fed-ISIC2019 \\ \cmidrule(l){1-6}
Centralized       & 0.615             &  & 0.957               &  & 0.614              \\
Local             & 0.494 $\pm 0.024$            &  & 0.785 $\pm 0.017$               &  & 0.290 $\pm 0.113 $             \\
SLViT   & 0.540 $\pm 0.029$           &  & 0.826 $\pm 0.018$              &  & 0.293 $\pm 0.133$             \\
SViBS (ours)              & 0.570 $\pm 0.011$                 &  & 0.836 $\pm 0.014$                  &  & 0.330 $\pm 0.042$                \\
FedAvg \cite{mcmahan2017communication}           & 0.564             &  & 0.894               &  & 0.476             \\
FedProx \cite{li2020federated}           & 0.568             &  & 0.892               &  & 0.472             \\
SCAFFOLD \cite{SCAFFOLD}           & 0.290            &  & 0.880               &  & 0.330             \\
MOON \cite{MOON}           & 0.570             &  & 0.903              &  & 0.450             \\
FeSTA \cite{visonVits}           & 0.638 $\pm$ 0.019            &  & 0.929 $\pm 0.001$              &  & 0.430 $\pm 0.048$            \\
FeSViBS (ours)           & \textbf{0.682 $\pm$ 0.021}    &  & \textbf{0.936$\pm$ 0.002}      &  & \textbf{0.534 $\pm$ 0.005}    \\ \bottomrule
\end{tabular}
\end{table}

\section{Results and Analysis}
Following \cite{flamby}, we used balanced accuracy in all experiments to evaluate the performance of the classification task across all datasets. This metric defines as the average recall on each class.
In Table. \ref{tab:my-table}, we compare the performance of FeSViBS and SViBS frameworks with other SOTA methods. FeSViBS consistently outperforms other methods on the three datasets with both IID and non-IID settings. More specifically, for HAM10000 (IID), FeSViBS outperforms all other methods with a $\textbf{4.4\%}$ gain in performance over FeSTA and approximately $\textbf{11\%}$ over FedAvg and FedProx ($\mu=0.006$). In the non-IID settings with both BloodMNIST and Fed-ISIC2019, FeSViBS maintains a high performance compared to other methods. Under extreme non-IID settings (Fed-ISIC2019), our approach demonstrated a performance improvement of $\textbf{10.4\%}$ compared to FeSTA and $\textbf{5.8\%}$ over FedAvg and FedProx, demonstrating the robustness of FeSViBS.

\begin{figure}[t!]
\centering{
\includegraphics[width=0.95\textwidth]{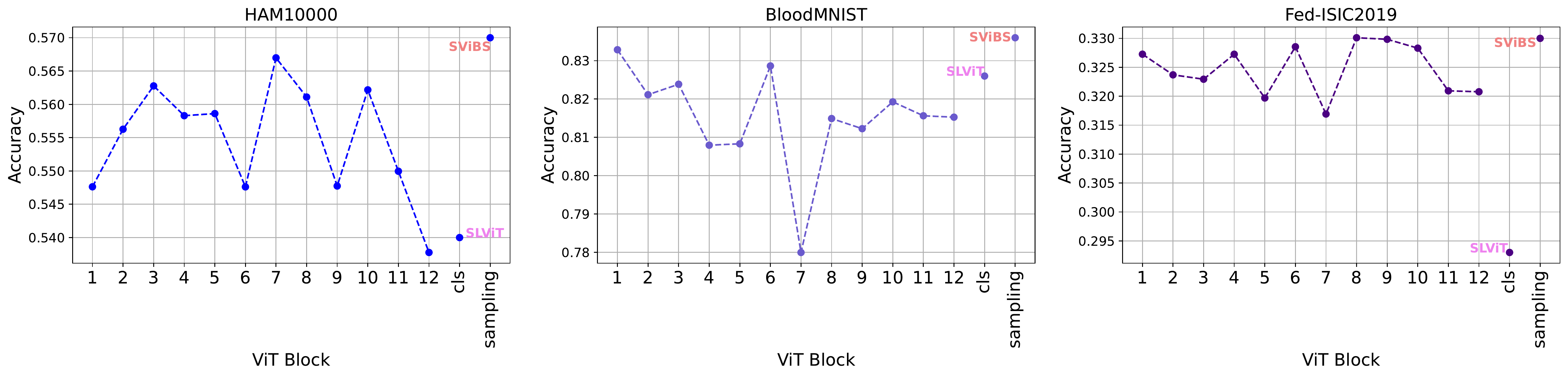}
}
\caption{Performance of each ViT block, sending \emph{cls} token (SLViT), and SViBS. Sampling from blocks 1 to 6 (SViBS) showed better performance than individual blocks.} 
\label{Figure: ablation blocks}
\end{figure}

We investigate the impact of sampling intermediate blocks in SViBS, by analysing the individual performance of intermediate features from specific blocks during training. The results in Fig. \ref{Figure: ablation blocks} demonstrate that the majority of individual blocks outperform the vanilla split learning setting (SLViT), which is dependent on the $cls$ token. On the other hand, SViBS shows dominant performance across datasets, where the sampling of ViT blocks provides augmented representations of the input images at different rounds, which improves the generalizability and then propagates this impact back to the ViT blocks and to the head. 

In a similar manner to FeSTA, we observe that the variance between the results achieved by different clients in FeSViBS is very low, especially when compared to SLViT. This shows that performance gain between clients is shared more or less equally, even in extreme non-IID settings, which is desirable.

\section{Ablation Study}

\begin{figure}[t!]
\centering{
\includegraphics[width=0.85\textwidth]{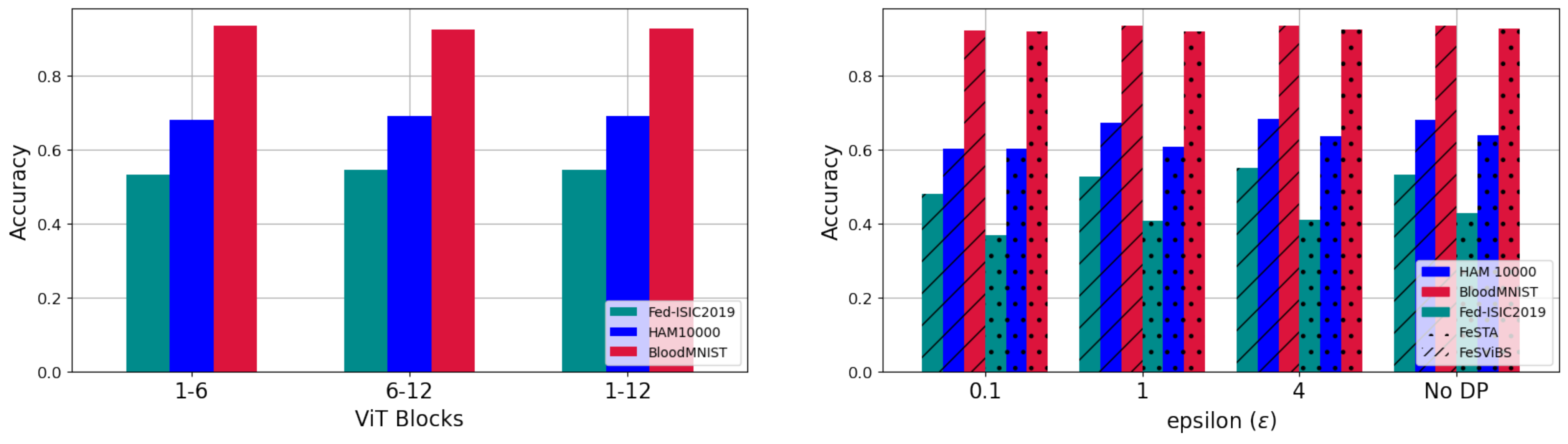}
}
\caption{FeSViBS performance with: \textbf{Left} different set of ViT blocks. \textbf{Right}: Differential Privacy with different $\epsilon$ values along with the original FeSViBS.} 
\label{Figure: ablation DP}
\end{figure}

\textbf{Set of ViT Blocks.} To study the impact of ViT blocks from which the intermediate features are sampled on the overall performance of FeSViBS, we carry out experiments choosing different sets of blocks. The results depicted in Fig. \ref{Figure: ablation DP} (left) show consistent performance for different sets of blocks across different datasets. This indicates that implementing FeSViBS with the first $6$ ViT blocks would reduce the computational cost without compromising performance. 

\subsubsection{\textbf{FeSViBS with Differential Privacy.}} Differential Privacy (DP) \cite{DP} is a widely-used approach that aims to improve the privacy of local client's data by adding noise. We conduct experiments where we add Gaussian noise to the client's head output ($h_c$). In such a scenario, DP makes it more challenging for a malicious/curious server to infer the client’s input from the smashed representations. With different $\epsilon$ values, the results in Fig. \ref{Figure: ablation DP} (right) show that FeSViBS maintains its performance even under a small $\epsilon $ value ($\epsilon = 0.1$), while also outperforming FeSTA under the same constraints.

\noindent \textbf{Number of Unifying Rounds.} We investigated the impact of reducing communication rounds (unifying rounds) on FeSViBS performance. However, our results showed that performance was maintained even with decreasing the number of communication rounds.

\section{Conclusion and Future Directions}
We proposed a novel Federated Split Learning of Vision Transformer with Block Sampling (FeSViBS), which utilizes FL, SL and sampling of ViT blocks to enhance the performance of the collaborative system. We evaluate FeSViBS framework under IID and non-IID settings on three real-world medical imaging datasets and demonstrate consistent performance. In the future, we aim to (i) extend our work and evaluate the privacy of FeSViBS under the presence of malicious clients/server, (ii) evaluate FeSViBS in the context of natural images and (iii) extend the current framework to multi-task settings. 

\bibliographystyle{splncs04}
\bibliography{paper1257}
\end{document}